\documentclass[letterpaper]{article}
\usepackage{aaai2026}  
\usepackage{times}     
\usepackage{helvet}
\usepackage{courier}
\usepackage[hyphens]{url}
\usepackage{graphicx}
\usepackage{natbib}
\usepackage{algorithm}   
\usepackage{algorithmicx} 
\usepackage{algpseudocode}

\usepackage{appendix}
\usepackage{array}
\usepackage{multirow} 

\usepackage{listings} 
\usepackage{xcolor}   
\usepackage{amsmath, amssymb}
\usepackage{cleveref}
\usepackage{booktabs}
\usepackage{caption}
\usepackage{multirow}  
\title{RoCo: Role-Based LLMs Collaboration for Automatic Heuristic Design}
\author {
    Jiawei Xu\textsuperscript{\rm 1},
    Fengfeng Wei\textsuperscript{\rm 2},
    Weineng Chen\textsuperscript{\rm 2},
}
\affiliations {
    \textsuperscript{\rm 1}School of Software Engineering, South China University of Technology, Guangzhou, China \\
    \textsuperscript{\rm 2}School of Computer Science and Engineering, South China University of Technology, Guangzhou, China\\
    ct20050202@mail.scut.edu.cn, fengfeng\_scut@163.com, cdchenwn@scut.edu.cn
}

\begin{document}

\maketitle

\begin{abstract}
Automatic Heuristic Design (AHD) has gained traction as a promising solution for solving combinatorial optimization problems (COPs). Large Language Models (LLMs) have emerged and become a promising approach to achieving AHD, but current LLM-based AHD research often only considers a single role. This paper proposes RoCo, a novel Multi-Agent Role-Based System, to enhance the diversity and quality of AHD through multi-role collaboration. RoCo coordinates four specialized LLM-guided agents—explorer, exploiter, critic, and integrator—to collaboratively generate high-quality heuristics. The explorer promotes long-term potential through creative, diversity-driven thinking, while the exploiter focuses on short-term improvements via conservative, efficiency-oriented refinements. The critic evaluates the effectiveness of each evolution step and provides targeted feedback and reflection. The integrator synthesizes proposals from the explorer and exploiter, balancing innovation and exploitation to drive overall progress. These agents interact in a structured multi-round process involving feedback, refinement, and elite mutations guided by both short-term and accumulated long-term reflections. We evaluate RoCo on five different COPs under both white-box and black-box settings. Experimental results demonstrate that RoCo achieves superior performance, consistently generating competitive heuristics that outperform existing methods including ReEvo and HSEvo, both in white-box and black-box scenarios. This role-based collaborative paradigm establishes a new standard for robust and high-performing AHD.
\end{abstract}

\section{Introduction}
Combinatorial optimization problems (COPs) underpin diverse real-world challenges, driving algorithmic innovation in areas such as path planning, electronic design automation (EDA), and manufacturing \citep{ma2018pathplanning, zebulum2018EDA, zhang2023reviewonmanufacturing}. Despite decades of advances in meta-heuristics, such as iterated local search, simulated annealing, and tabu search, for solving NP-hard combinatorial optimization problems \citep{Mart2018Handbook}, such design remains a major bottleneck. This process often demands significant domain-specific expertise and extensive trial-and-error, particularly in applications like job scheduling, logistics planning, and robotics \citep{Tan2021}. To address this challenge, Automatic Heuristic Design (AHD) has emerged as a compelling research direction. It seeks to evolve or compose heuristics automatically, reducing reliance on expert-crafted rules. Notable approaches include Genetic Programming (GP)-based methods\citep{zhang2023survey} and Hyper-Heuristics (HHs) \citep{pillay2018hyper}, which operate by searching within predefined heuristic or operator spaces. Recent efforts have explored learning-based improvement heuristics using deep reinforcement learning for routing problems, showing better performance than traditional hand-crafted rules \citep{wu2021drl}. However, GP- and HH-based methods remain limited by their reliance on human-defined heuristic spaces, making them less adaptable to novel or black-box tasks. These limitations have driven a new wave of research into more expressive, flexible, and generalizable approaches to automatic heuristic generation.

In recent years, large language models (LLMs), such as GPT-4 \citep{achiam2023gpt}, have demonstrated impressive capabilities across various reasoning tasks \citep{hadi2023survey}, leading to the development of LLM-based automatic heuristic design (AHD) methods that iteratively refine heuristics through evolutionary program search \citep{liu2024large, chen2023evoprompting}. Early works such as FunSearch \citep{romera2024funsarch} and EoH \citep{liu2024eoh} adopt population-based frameworks that evolve heuristic functions using LLMs within fixed algorithmic templates. ReEvo \citep{ye2024reevo} enhances heuristic quality through reflection-based reasoning \citep{shinn2023reflexion}, while HSEvo \citep{dat2025hsevo} introduces diversity metrics and harmony search to improve population diversity. Recent work MCTS-AHD\citep{zheng2025monte} extends this line of research by organizing heuristics in a tree-based structure guided by Monte Carlo Tree Search, enabling structured refinement beyond conventional population-based methods. Together, these methods represent a growing family of LLM-EPS techniques that automate the design of high-quality heuristics for combinatorial and black-box optimization tasks.

Although recent LLM-based approaches to automatic heuristic design (AHD) have demonstrated success in evolving algorithms via population-level or iterative sampling methods, their performance remains unstable when applied to combinatorial optimization problems (COPs), especially in black-box scenarios. These approaches generally employ a single LLM with various prompt strategies, but do not explicitly model distinct agent roles or structured inter-agent coordination. Another relevant effort, LEO \citep{brahmachary2025LEO}, introduces separate explore and exploit pools to balance diversity and refinement in LLM-driven optimization. However, it lacks explicit role modeling or inter-agent communication, and its performance on complex COPs remains inconsistent. This ultimately limits their ability to adaptively decompose complex tasks or exploit diverse reasoning patterns.

Out of the above consideration, our work draws inspiration from LLM-based Multi-Agent Systems (MAS) \citep{zhang2023exploringllmagents, li2024MASsurvey}, introducing a collaborative agent framework, named RoCo(\textbf{Ro}le-based LLMs \textbf{Co}llaboration), in which LLM-based agents assume specialized roles, such as explorer, exploiter, critic, and integrator, to collectively generate, critique, and refine algorithmic heuristics. Our method extends the population-based architecture of EoH, inheriting its advantages in maintaining heuristic diversity and leveraging population-level selection, while introducing an additional layer of role specialization to enable structured agent-level cooperation. Exploration and exploitation agents generate individuals, critics offer reflection and suggestions, and integrators consolidate the individual outputs of explorers and exploiters to generate new and improved heuristics, balancing exploration and exploitation. This design of critic and integrator is further inspired by multi-agent debate systems \citep{du2023multiagentdebate, liang2023multiagentdebate}, where dialogic reasoning and reflective interaction support deep thinking and innovation. However, unlike adversarial debates, our agents engage in cooperative self-improvement, guided by critic feedback, and equipped with short-term (per round) and long-term (cross-round) reflection mechanisms. The long-horizon reflection mechanism is distilled from short-term feedback across rounds, capturing both successful patterns and failed heuristic attempts. Inspired by recent work that highlights the value of learning from exploration failures \citep{an2023learningfrommistakes, song2024trial}, it enables agents to accumulate insight from both progress and missteps. Such a design leverages both the coordination potential of multi-agent collaboration and the iterative divergence–convergence dynamics of debate, fostering more robust and generalizable heuristic discovery.

In experiments, we apply our approach, RoCo, to five combinatorial optimization problems (COPs) under both white-box and black-box settings. RoCo consistently accelerates the convergence of heuristic search in white-box scenarios and enhances performance stability in black-box scenarios by evolving high-quality heuristics across diverse problem structures.

Our main contributions are as follows: (1) We propose RoCo, a novel LLM-based automatic heuristic design (AHD) framework that introduces role-specialized agents:explorers, exploiters, critics, and integrators, for structured collaboration in generating and refining heuristics. (2) RoCo incorporates coordinated multi-agent interaction and a reflection process that captures both successful strategies and failed heuristic attempts, enabling agents to learn adaptively from mistakes and improve search quality over time.

\section{Background}

\subsection{COP Formulation}
A combinatorial optimization problem (COP) instance is defined as \citep{zheng2025monte}: 
\begin{equation}
    P = (I_P, S_P, f) \label{eq:cop-definition}
\end{equation}

where:
\begin{itemize}
    \item $I_P$: the set of input instances (e.g., coordinates in TSP),
    \item $S_P$: the set of feasible solutions,
    \item $f : S_P \to \mathbb{R}$: the objective function to minimize.
\end{itemize}

A heuristic $h : I_P \to S_P$ maps an input to a feasible solution. The goal of Automatic Heuristic Design (AHD) is to find the best heuristic $h^*$ in a heuristic space $\mathcal{H}$, maximizing the expected performance:
\begin{equation}
    h^* = \arg\max_{h \in \mathcal{H}} g(h), \quad g(h) = \mathbb{E}_{ins \sim D}\left[ -f(h(ins)) \right] \label{eq:ahd-objective}
\end{equation}

\subsection{LLMs for AHD} LLM-based Evolutionary Program Search (LLM-EPS) integrates large language models into evolutionary computation frameworks to automate the design of heuristics. Representative methods include FunSearch \citep{romera2024funsarch}, which adopts an island-based strategy for evolving heuristics in mathematical problems; EoH \citep{liu2024eoh}, which applies genetic algorithms with chain-of-thought prompting to improve solutions for tasks like TSP and bin packing ; and ReEvo \citep{ye2024reevo}, which introduces reflective evolution using paired LLMs. More recent approaches like HSEvo \citep{dat2025hsevo} and MEoh \citep{yao2025MEoh} explore diversity-driven harmony search with genetic algorithms and multi-objective strategies with dominance-dissimilarity mechanisms, respectively. MCTS-AHD \citep{zheng2025monte} mitigates this by using a tree-based structure to guide LLM exploration more flexibly, while RedAHD \citep{thach2025redahd} further enables end-to-end heuristic design by prompting LLMs to construct and refine problem reductions without relying on fixed frameworks. To support the implementation and evaluation of such LLM-EPS methods, LLM4AD has been developed as a unified Python platform, integrating modularized blocks for search methods, algorithm design tasks, and LLM interfaces, along with a unified evaluation sandbox and comprehensive support resources \citep{liu2024llm4ad}.

\subsection{LLMs-based Multi-Agent System}  Recent work on LLM-based Multi-Agent Systems (MAS) has increasingly focused on collaboration mechanisms \citep{tran2025Multiagentcollaborationmechanisms, li2023theoryformas, pan2024agentcoord} that facilitate interaction among agents, allowing them to coordinate actions, share information, and solve tasks collectively. Early methods explore fixed interaction patterns such as debate to enhance reasoning \citep{du2023improve_factuality, liang2023multiagentdebate}, while later approaches introduce adaptive communication protocols for multi-round decision-making \citep{liu2023dynamic}. Other frameworks support broader applications by organizing LLM agents to work jointly on complex problems \citep{hong2023metagpt, wu2024autogen}.Our method follows a role-based collaboration protocol \citep{chen2023agentverse, talebirad2023multiagent}, where agents operate under distinct predefined roles to tackle subgoals modularly and cooperatively \citep{tran2025Multiagentcollaborationmechanisms}.

\section{The Proposed RoCo}

\subsection{Role-Based LLM Agent System Definition}
To enable effective and scalable AHD, we propose a role-based collaborative system (RoCo) powered by multiple large language model (LLM) agents. This system leverages structured collaboration through clearly defined agent roles, shared memory, and multi-round interaction. Formally, our system is defined as \citep{tran2025Multiagentcollaborationmechanisms}: 
\begin{equation}
\mathcal{S} = (\mathcal{A}, \mathcal{R}, \mathcal{E}, \mathcal{C}, x_{\text{collab}}, y_{\text{collab}}) \label{eq:system-definition}
\end{equation}
\begin{itemize}
    \item $\mathcal{A} = \{a_i\}_{i=1}^n$: a set of $n$ LLM-based agents.
    \item $\mathcal{R} = \{\text{explorer}, \text{exploiter}, \text{critic}, \text{integrator}\}$: the predefined set of agent roles.
    \item $\mathcal{E}$: the shared environment (e.g., population pool, reflection history).
    \item $\mathcal{C}$: a set of communication channels among agents.
    \item $x_{\text{collab}}$: the system input (e.g., selected individuals from EoH population).
    \item $y_{\text{collab}}$: the system output (e.g., improved heuristics via collaboration and reflection).
\end{itemize}

Each agent $a_i$ is further characterized by the tuple:
\begin{equation}
a_i = (L_i, R_i, S_i, T_i) \label{eq:agent-characterization}
\end{equation}
\begin{itemize}
    \item $L_i$: the underlying language model (e.g., GPT-4).
    \item $R_i \in \mathcal{R}$: the agent's role.
    \item $S_i$: agent state, including current individuals and reflection buffer.
    \item $T_i$: tools (e.g., Python interpreter).
\end{itemize}

Finally, the entire system executes collaborative optimization as a function over its inputs and shared environment:
\begin{equation}
y_{\text{collab}} = \mathcal{S}(\mathcal{O}_{\text{collab}}, \mathcal{E}, x_{\text{collab}} \mid \mathcal{A}, \mathcal{C}) \label{eq:collab-optimization}
\end{equation}

\subsection{Heuristic Representation}

The heuristic representation in RoCo follows the structure of the EoH framework. Each heuristic $h \in \mathcal{H}$ is represented as a tuple:

\begin{enumerate}
    \item \textbf{A natural language description} $\mathsf{desc}(h)$ outlining the core idea of $h$.
    \item \textbf{A code implementation} $\mathsf{code}(h)$, defined as a Python function that maps an input instance $ins \in I_P$ to a feasible solution $h(ins) \in S_P$.
    \item \textbf{A fitness score} $g(h)$, measuring the expected performance of the heuristic across a distribution of instances, as defined in Equation~\ref{eq:ahd-objective}.
\end{enumerate}

This structured representation enables seamless integration of newly generated heuristics into the EoH evaluation pipeline, while ensuring compatibility with standard analysis and benchmarking tools.

\subsection{System Overview of the RoCo framework}

\begin{figure*}[t]
    \centering
    \includegraphics[width=0.88\textwidth]{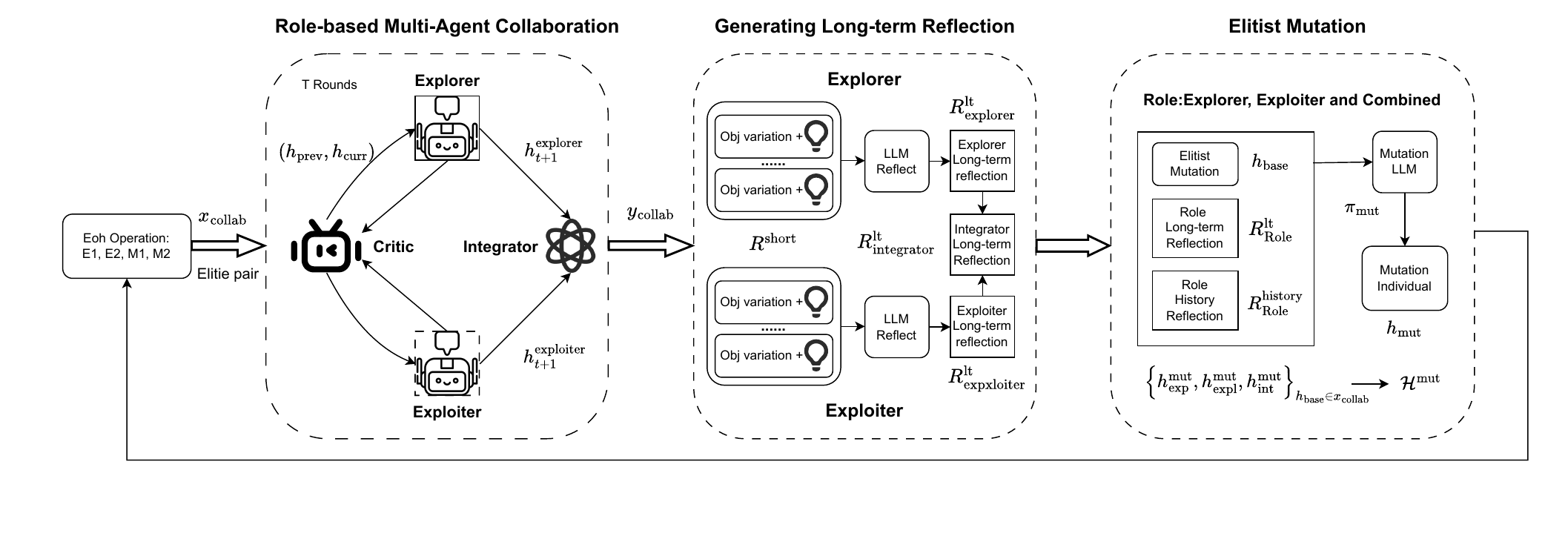}
    \caption{The architecture of the RoCo system, which integrates multi-agent collaboration into the evolutionary heuristic generation process.}
    \label{fig:roco-arch}
\end{figure*}

Our proposed RoCo system is a multi-agent collaborate-reflection module designed to enhance heuristic generation in the evolutionary framework. As illustrated in Figure~\ref{fig:roco-arch}, RoCo operates as a plug-in system integrated within the Evolution of Heuristics (EoH) paradigm, extending the standard evolutionary loop with collaborative reasoning, structured role division, and multi-agent learning.

At each generation \( g \), the EoH framework maintains a population \( P^g = \{ h_1^g, h_2^g, \ldots, h_N^g \} \) of heuristics. The population is evolved by applying four prompt-based operators, classified into two categories:

\begin{itemize}
    \item \textbf{Exploration Operators (E1, E2)}: generate heuristics by introducing diversity or combining conceptual patterns from multiple parents.
    \item \textbf{Modification Operators (M1, M2)}: refine or adjust existing heuristics via mutation or parameter tuning. The details of Eoh operators are exhibited in Appendix \ref{eohdetails}.
\end{itemize}

This results in a candidate pool of up to \( 4N \) new heuristics. Among these, a subset of heuristics is selected and passed to our multi-agent collaboration system \( \mathcal{S} \) to undergo a structured refinement process.

Formally, two heuristics \( h^{(1)} \) and \( h^{(2)} \) are sampled from \( P^g \) and passed as input:

\begin{equation}
x_{\text{collab}} = \left( h^{(1)}, h^{(2)} \right) \in P^g
\end{equation}
The selection of elite individuals follows the strategy described in Appendix \ref{eohdetails}.

The RoCo system then initiates a multi-round collaboration involving agents with distinct functional roles. The output is a set of improved heuristics directly produced from multi-round collaboration, covering different reasoning strategies:

\begin{equation}
y_{\text{collab}} = \left\{ h_{g+1}^{\text{explorer}}, h_{g+1}^{\text{exploiter}}, h_{g+1}^{\text{integrator}}, \ldots \right\}
\end{equation}

To construct the next-generation population \( P^{g+1} \), we first gather all heuristics generated by RoCo in generation \( g \), including the collaborative outputs \( y_{\text{collab}} \) produced during debate and the full set of elite-mutated individuals \( \mathcal{H}^{\text{mut}} \). These together form the RoCo candidate set:
\begin{equation}
    \mathcal{C}^{g+1}_{\text{RoCo}} = y_{\text{collab}} \cup \mathcal{H}^{\text{mut}}
\end{equation}

We then merge this set with the heuristic pool \( P_{\text{eoh}}^{g+1} \) generated by standard EoH operators to obtain the full candidate set:
\begin{equation}
    \mathcal{C}^{g+1}_{\text{full}} = \mathcal{C}^{g+1}_{\text{RoCo}} \cup P_{\text{eoh}}^{g+1}
\end{equation}

From this combined set, we deterministically select the top \( N \) individuals with the best objective values to form the next population:
\begin{equation}
    P^{g+1} = \text{TopN} \left( \mathcal{C}^{g+1}_{\text{full}}, N \right)
\end{equation}

This approach ensures that high-quality heuristics, whether discovered through structured collaboration, memory-driven mutation, or standard evolutionary operators, are retained in the population. The details are shown in Algorithm \ref{alg:roco}.

\subsection{Role Specialization and Agent Design}
Each agent in the RoCo system is assigned a specific role to facilitate decomposition of the overall heuristic optimization task into well-defined sub-tasks. The four main roles are:

\textbf{Explorer}  
Responsible for generating diverse heuristics through conceptual exploration:  
\begin{equation}
h_{t+1}^{\text{explorer}} = \pi_{\text{explorer}} \left( h_t, f_t^{\text{cri}}\right)
\end{equation}

\textbf{Exploiter}  
Focuses on local improvement by refining promising candidates:  
\begin{equation}
h_{t+1}^{\text{exploiter}} = \pi_{\text{exploiter}} \left( h_t, f_t^{\text{cri}} \right)
\end{equation}

\textbf{Critic}  
Acts as an evaluator and reflective thinker, identifying strengths and limitations in heuristics and providing structured feedback. Evaluates pair \( (h_{\text{prev}}, h_{\text{curr}}) \) and gives feedback and reflection:  
\begin{equation}
\label{critic_equ}
    \left( f_t^{\text{cri}}, r_t^{\text{cri}} \right) = \pi_{\text{critic}} \left( h_{\text{prev}}, h_{\text{curr}} \right)
\end{equation}

\textbf{Integrator}  
Fuses two candidates with both exploration and exploitation based on their objective scores:  
\begin{equation}
    h^{\text{int}} = \pi_{\text{integrator}} \left( h^{\text{explorer}}, h^{\text{exploiter}} \right)
\end{equation}
After all collaborative rounds, it performs \emph{elitist fusion} of the best-performing explorer and exploiter heuristics. Details are exhibited in Algorithm \ref{alg:roco}

Each agent \( a_i = (L_i, R_i, S_i, T_i) \) operates with its own language model backend \( L_i \) (e.g., GPT-4), a predefined role \( R_i \in \mathcal{R} \) such as explorer that governs its behavioral policy, an internal state \( S_i \) containing its current candidate heuristics and optional reflection buffers, and a toolset \( T_i \) that enables it to evaluate heuristics by executing them and obtaining objective values. The core behavior of each agent is determined by a role-specific policy function \( \pi_{\text{role}} \), which guides its reasoning and interactions within a multi-round collaborative loop.

\subsection{Collaboration and Reflection Mechanisms}
RoCo introduces a multi-round collaborative debate process where agents iteratively refine heuristics over \( T \) steps. In each round, the critic agent plays a central role by evaluating the most recent heuristic pair and providing feedback, as described in Equation~\ref{critic_equ}.

After \( T \) rounds, the accumulated critic reflections \( R_{\text{role}}^{\text{short}} \) are synthesized into long-term role-specific memory by incorporating not only the raw feedback but also the associated objective values and their changes:

\begin{equation}
   R_{\text{role}}^{\text{lt}} = \text{LTReflect}(R_{\text{role}}^{\text{short}}, g_{t-1}, g_t, \Delta g_t)
\end{equation}

Here, $g_{t-1}$ and $g_t$ are the objective values before and after the $t$-th reflection, with $\Delta g_t = g_t - g_{t-1}$ capturing per-round performance changes. This helps the model infer useful patterns by linking reflection to performance dynamics.

To enable continual learning, each role maintains its own historical memory \( R_{\text{role}}^{\text{history}} \), formed by aggregating long-term reflections from previous generations. These role-specific historical reflections provide learning signals beyond the current population and enable richer generalization across diverse search states.

Additionally, for integrative reasoning, the long-term reflections from explorer and exploiter can be merged to support integrative generation strategies:
\begin{equation}
   R_{\text{integrator}}^{\text{lt}} = \text{Merge}\left(R_{\text{explorer}}^{\text{lt}}, R_{\text{exploiter}}^{\text{lt}}\right) 
\end{equation}

To complement collaboration-based generation, RoCo supports a memory-guided elite mutation strategy. For each elite individual \( h_{\text{base}} \), three types of memory-augmented mutations are performed: 
\begin{equation}
    \begin{aligned}
    h^{\text{mut}}_{\text{exp}} &= \pi_{\text{mutation}}^{\text{exp}}(h_{\text{base}}, R_{\text{explorer}}^{\text{lt}}, R_{\text{explorer}}^{\text{history}}) \\
    h^{\text{mut}}_{\text{expl}} &= \pi_{\text{mutation}}^{\text{expl}}(h_{\text{base}}, R_{\text{exploiter}}^{\text{lt}}, R_{\text{exploiter}}^{\text{history}}) \\
    h^{\text{mut}}_{\text{int}} &= \pi_{\text{mutation}}^{\text{int}}(h_{\text{base}}, R_{\text{integrator}}^{\text{lt}}, R_{\text{integrator}}^{\text{history}})
\end{aligned}
\end{equation}

All resulting mutated individuals from all elite bases are collected into a unified set:
\begin{equation}
    \mathcal{H}^{\text{mut}} = \left\{ h^{\text{mut}}_{\text{exp}}, h^{\text{mut}}_{\text{expl}}, h^{\text{mut}}_{\text{int}} \right\}_{h_{\text{base}} \in x_{\text{collab}}}
\end{equation}

This mechanism mimics a memory-augmented exploitation strategy and supports lifelong learning across generations through diverse role-informed mutations.

\begin{figure*}[t] 
    \centering
    \includegraphics[width=1.0\textwidth]{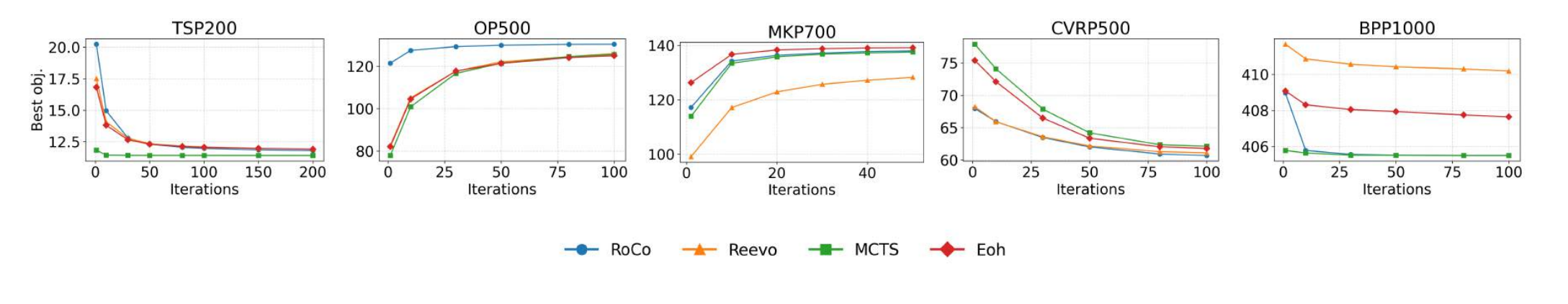} 
    \caption{Evolution curves of different LLM-based AHD methods, plotting the all-time best objective value w.r.t. the number of solution evaluations. The curves, for five datasets (TSP, MKP, OP, BPP, CVRP) each with 1 scales, are averaged over 3 runs with small variances observed.}
    \label{fig:whiteboxcurve}
\end{figure*}

\section{Experiments}

This section first introduces the experimental setup, including baseline algorithms and implementation details. We then evaluate the proposed method under two metaheuristic frameworks: Ant Colony Optimization (ACO) \citep{dorigo2007aco} and Guided Local Search (GLS) \citep{arnold2019gls}, covering a range of combinatorial optimization problems (COPs). Under the ACO framework, we conduct both white-box and black-box evaluations to assess the method’s generalization and robustness. In the GLS framework, we further demonstrate the effectiveness of our approach on the Traveling Salesman Problem instances. Finally, ablation studies are presented to analyze the contribution of each system component.

\subsection{Experimental Setup}
\label{sec:exp_setup}
\textbf{Benchmarks:} To evaluate the generality and robustness of our approach, we conduct experiments across a suite of combinatorial optimization problems (COPs) spanning multiple structural categories. Specifically, we include: the Traveling Salesman Problem (TSP), Capacitated Vehicle Routing Problem (CVRP), and Orienteering Problem (OP) as representative routing tasks; the Multiple Knapsack Problem (MKP) for subset selection; and the Bin Packing Problem (BPP) for grouping-based challenges. These benchmarks are widely adopted in the COP literature and collectively cover a broad spectrum of algorithmic patterns and difficulty profiles.

\textbf{Baselines:} To assess the effectiveness of our system in generating high-quality heuristics, we compare it against a set of strong baselines. These include several representative LLM-based automatic heuristic design (AHD) approaches such as EoH, ReEvo, and HSEvo, as well as the most recent MCTS-AHD framework. Additionally, we include DeepACO \citep{ye2023deepaco}, a neural combinatorial optimization method tailored for ACO frameworks, along with manually designed heuristics, ACO \citep{dorigo2007aco}.

\textbf{Settings:} In our RoCo multi-agent system, the number of collaboration rounds is fixed to 3, with a total LLM API budget capped at 400 calls per generation. The underlying EoH framework uses a population size of 10. All experiments are conducted using GPT-4o-mini as the language model. Role-specific temperatures are set as follows: explorer with 1.3, exploiter with 0.8, and default 1.0 for all other roles. Detailed dataset configurations and general framework settings are provided in Appendix \ref{exp_details}.

\begin{table*}[h]
\centering
\caption{Designing heuristics with the ACO general framework for solving TSP, OP, CVRP, MKP, and offline BPP under the white-box setting. Each test set contains 64 instances and LLM-based AHD methods' performances are averaged over three runs.}
\label{tab:whitebox}
\scriptsize  
\setlength{\tabcolsep}{1.0mm}  
\renewcommand{\arraystretch}{1.1}  
\begin{tabular}{lccc|ccc|ccc|ccc|ccc}  
\hline
\multirow{2}{*}{\textbf{Methods}} 
& \multicolumn{3}{c|}{\textbf{TSP Obj.↓}} 
& \multicolumn{3}{c|}{\textbf{OP Obj.↑}} 
& \multicolumn{3}{c|}{\textbf{CVRP Obj.↓}} 
& \multicolumn{3}{c|}{\textbf{MKP Obj.↑}} 
& \multicolumn{3}{c}{\textbf{Offline BPP Obj.↓}} \\ 
\cline{2-16}  
& N=50 & N=100 & N=200  
& N=50 & N=100 & N=200  
& N=50 & N=100 & N=200  
& N=100 & N=200 & N=500 
& N=500 & N=700 & N=1000 \\ 
\hline
ACO               &6.064 &9.042&13.555  & 14.981 & 29.979 & 53.380 &11.623 &19.131&35.441 &21.673 &42.802 &102.320  &209.132 &291.436 &416.412  \\
DeepACO           &5.811 &8.186&11.615  & 14.574 & 30.474 & 55.458 &9.130  &15.242&\textbf{27.057}&21.958 &43.170 &102.975 &203.125 &286.685  &406.822  \\
\hline
\multicolumn{16}{c}{LLM-based AHD: GPT-4o-mini} \\ \hline
Eoh     &5.830 &8.247 &11.771 & 15.284 & 31.242 & 56.320 &9.379&15.642&27.801&23.141 &41.952 &101.472 &204.578 &218.516 &408.672   \\
ReEvo   &5.770 &8.091 &11.619 & 15.267 & 31.161 & 56.380 &9.095&15.570&27.723&23.039 &41.576 &98.932   &205.141 &287.125  &409.531  \\
HsEvo   &5.830 &8.315 &12.058 & 14.910 & 30.414 & 54.791 &9.589&16.426&29.499&\textbf{23.277} &42.454 &102.993  &205.187 &287.063  &409.500  \\
MCTS-AHD&5.790 &\textbf{8.052} &\textbf{11.380} & \textbf{15.652} & 31.428 & 56.205 &9.369&15.938&28.375&23.207 &42.353 &103.110  &202.906 &283.797 &\textbf{404.734} \\
RoCo(Ours) &\textbf{5.766} &8.115 &11.672 & 15.586 & \textbf{31.653} & \textbf{57.365}&\textbf{8.966}&\textbf{15.148} &27.262 &23.249 &\textbf{42.479} &\textbf{103.580} &\textbf{202.891} &\textbf{283.734} &404.766  \\
\hline  
\end{tabular}
\end{table*}

\begin {table*}[h]
\centering
\caption {Designing heuristics with the ACO general framework for solving TSP, OP, CVRP, MKP, and offline BPP under the black-box prompt setting. Each test set contains 64 instances and LLM-based AHD methods' performances are averaged over three runs.}
\label {tab:black_box_table}
\scriptsize 
\setlength {\tabcolsep}{1.0mm} 
\renewcommand {\arraystretch}{1.1} 
\begin {tabular}{lccc|ccc|ccc|ccc|ccc}
\hline
\multirow{2}{*}{\textbf{Methods}} 
& \multicolumn {3}{c|}{\textbf {TSP Obj.↓}}
& \multicolumn {3}{c|}{\textbf {OP Obj.↑}}
& \multicolumn {3}{c|}{\textbf {CVRP Obj.↓}}
& \multicolumn {3}{c|}{\textbf {MKP Obj.↑}}
& \multicolumn {3}{c}{\textbf {Offline BPP Obj.↓}} \\
\cline {2-16}
& N=50 & N=100 & N=200
& N=50 & N=100 & N=200 
& N=50 & N=100 & N=200 
& N=100 & N=200 & N=500 
& N=500 & N=700 & N=1000 \\ 
\hline
\multicolumn{16}{c}{LLM-based AHD: GPT-4o-mini} \\ \hline
Eoh         &5.823 &8.263 &11.885 &15.250 &31.162 &56.364 &9.321 &15.732 &27.881 &23.137    &41.943 &101.556 &204.313 &285.859 &407.719   \\
ReEvo       &\textbf{5.767} &\textbf{8.096} &\textbf{11.597} &15.238 &\textbf{31.290} &56.643 &9.472 &16.293 &28.983 &\textbf{23.318}  &\textbf{42.719} &104.236 &204.203 &285.906 &407.891 \\
HsEvo       &5.826 &8.286 &11.867 &15.205 &30.589 &51.874 &9.849 &17.455 &31.393 &23.155  &42.022 &101.820 &\textbf{204.203} &285.844 &\textbf{407.672} \\
MCTS-AHD    &5.834 &8.243 &11.785 &15.038 &30.882 &50.887 &9.293 &\textbf{15.683} &27.805 &23.269  &42.530 &103.781 &204.281 &285.953 &407.781 \\
RoCo (Ours) &5.837 &8.244 &11.763 &\textbf{15.309} &31.239 &\textbf{56.716} &\textbf{9.283} &15.706 &\textbf{27.782} &23.297  &42.637 &\textbf{104.241} &204.312 &\textbf{285.734} &407.688 \\
\hline
\end{tabular}
\end{table*}

\subsection{Main Results}
\label{sec:main}

\textbf{Heuristic Measures for Ant Colony Optimization(ACO).} We first evaluate our RoCo-based heuristic generation system under the Ant Colony Optimization (ACO) framework, a widely adopted paradigm that integrates stochastic solution sampling with pheromone-guided search. Within this framework, our method focuses on designing heuristic functions that estimate the potential of solution components, which are then used to bias the construction of solutions across iterations. We apply this approach to five classic and diverse NP-hard combinatorial optimization problems: TSP, CVRP, OP, MKP, and offline BPP. 

Under the white-box prompt setting , we evaluate LLM-based Automatic Heuristic Design (AHD) methods across five COPs. As shown in Table ~\ref{tab:whitebox}, experiments conducted on 64 instances per problem demonstrate that our RoCo method achieves competitive performance. Notably, RoCo outperforms traditional ACO across all problems and surpasses DeepACO in most cases. Compared to other LLM-AHD approaches, RoCo attains the highest scores in 10 out of 15 problem-size combinations, demonstrating its effectiveness for heuristic design in white-box optimization.

As visualized in Figure \ref{fig:whiteboxcurve}, which plots the evolution of all-time best objective values across diverse problem types and scales, RoCo not only achieves strong final performance but also demonstrates faster convergence across most problem types beyond TSP, where all methods converge similarly. Specifically, in CVRP, MKP, BPP, and OP, RoCo reaches near-optimal performance within fewer iterations compared to baselines, indicating superior sample efficiency. Furthermore, RoCo exhibits robust scalability across problem sizes, consistently maintaining leading or near-leading performance as the instance scale increases—from small to large sizes—across all five combinatorial optimization problems. This highlights RoCo's capacity to generalize its heuristic design abilities across both simple and complex scenarios under white-box prompting conditions.

\begin{figure*}[t] 
    \centering
    \includegraphics[width=1.0\textwidth]{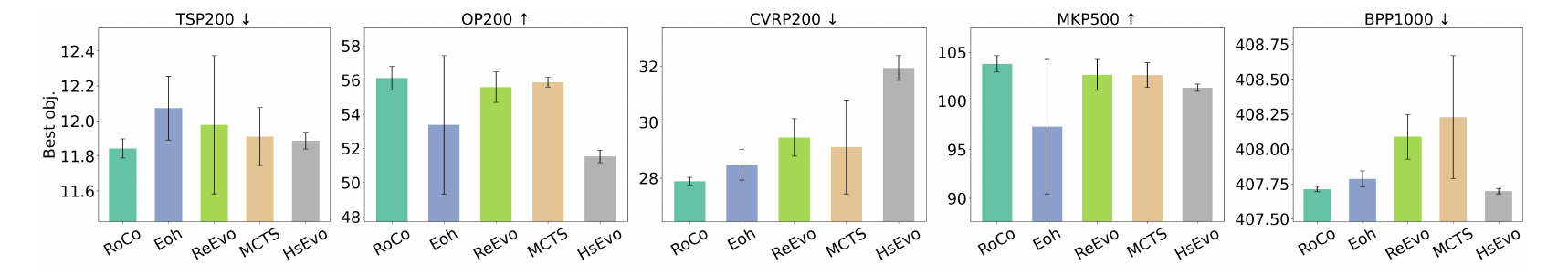} 
    \caption{Heuristic performance evaluation (mean and standard deviation) across diverse datasets (TSP200, OP200, CVRP200, MKP500, BPP1000) under black-box setting, with results aggregated from four independent runs per dataset, using LLM-based AHD.}
    \label{fig:blackbox}
\end{figure*}

Under the black-box setting, where LLMs receive only the textual prompt without access to the internal solver state, we evaluate five AHD methods across TSP, OP, CVRP, MKP, and offline BPP. As shown in Table~\ref{tab:black_box_table}, RoCo achieves strong and stable performance across all problem types and sizes. Notably, RoCo yields the best average results in numerous cases, and performs on par with the best-performing method (ReEvo) in terms of overall heuristic quality. Complementing these results, Figure~\ref{fig:blackbox} illustrates the distribution of objective values across multiple runs. RoCo consistently demonstrates smaller standard deviations, indicating enhanced robustness and stability under prompt-only settings. These findings highlight the generality and reliability of RoCo in automatic heuristic design, even under limited-access black-box conditions.

\begin{table}[t]
\setlength{\tabcolsep}{3.5pt}  
\centering
\caption{Ablation study of RoCo components on TSP with white-box and black-box prompting.}
\label{tab:ablation}
\begin{tabular}{lcc}
\toprule
\textbf{Method} & \textbf{White box} $\downarrow$ & \textbf{Black box} $\downarrow$ \\
\midrule
w/o Explorer       & 8.341 ± 0.043 & 8.269 ± 0.005 \\
w/o Exploiter      & 8.286 ± 0.049 & 8.400 ± 0.103 \\
w/o Integrator     & 8.265 ± 0.039 & 8.641 ± 0.428 \\
w/o Elite Mutation & 8.381 ± 0.080 & 8.266 ± 0.016 \\
w/o MAS            & 8.293 ± 0.062 & 8.263 ± 0.019 \\
\midrule
Collaborate Rounds = 1         & 8.333 ± 0.079 & 9.341 ± 1.428 \\
Collaborate Rounds = 2         & 8.289 ± 0.067 & 8.608 ± 0.512 \\
Collaborate Rounds = 3         & 8.261 ± 0.010 & 8.254 ± 0.017 \\
Collaborate Rounds = 4         & 8.275 ± 0.039 & 8.277 ± 0.026 \\
Collaborate Rounds = 5         & 8.267 ± 0.036 & 8.280 ± 0.027 \\
\midrule
RoCo (full)        & \textbf{8.256 ± 0.023} & \textbf{8.256 ± 0.014} \\
EoH (baseline)     & 8.257 ± 0.027 & 8.327 ± 0.064 \\
\bottomrule
\end{tabular}
\end{table}

\begin{table*}[htbp]
\centering
\caption{Evaluation results of different local search (LS) variants in terms of optimality gaps.}
\label{tab:gls}
\setlength{\tabcolsep}{0mm} 
\renewcommand{\arraystretch}{1.2} 
\footnotesize
\begin{tabular}{@{}p{3.5cm}p{2.2cm}
                >{\centering\arraybackslash}p{2.1cm}
                >{\centering\arraybackslash}p{2.1cm}
                >{\centering\arraybackslash}p{2.1cm}
                >{\centering\arraybackslash}p{2.1cm}@{}}
\toprule
\textbf{Method} & \textbf{Type} & \textbf{TSP20} & \textbf{TSP50} & \textbf{TSP100} & \textbf{TSP200} \\
               &               & Opt. gap (\%)  & Opt. gap (\%)  & Opt. gap (\%)   & Opt. gap (\%) \\
\midrule
NeuOpt$^*$            & LS+RL    & 0.000 & 0.000 & 0.027 & 0.403 \\
GNNGLS                & GLS+SL   & 0.000 & 0.052 & 0.705 & 3.522 \\
EoH \citep{liu2024eoh}                  & GLS+LHH  & 0.000 & 0.000 & 0.025 & 0.338 \\
\midrule
KGLS           & GLS      & 0.004 & 0.017 & 0.002 & 0.284 \\
KGLS-ReEvo    & GLS+LHH  & 0.000 & 0.000 & \textbf{0.000} & 0.216 \\
KGLS-MCTS-AHD      & GLS+LHH  & 0.000 & 0.000 & 0.001 & 0.214     \\
KGLS-RoCo         & GLS+LHH  & 0.000 & 0.018 & 0.001 & \textbf{0.188} \\
\bottomrule
\multicolumn{6}{l}{\footnotesize $^\S$ Results of the first three rows (NeuOpt$^*$, GNNGLS, EoH) are derived from ReEvo \citep{ye2024reevo}.} \\
\end{tabular}
\end{table*}

\textbf{Guided Local Search (GLS).} We further evaluate RoCo in the GLS framework by evolving penalty heuristics that guide local search in escaping local optima. Specifically, we embed the best heuristics produced by RoCo into KGLS, a state-of-the-art GLS baseline, and compare its performance (denoted as KGLS-RoCo) against several competitive methods, including NeuOpt \citep{ma2023neuopt}, GNNGLS \citep{hudson2021gnngls}, EoH \citep{liu2024eoh}, and KGLS-ReEvo \citep{arnold2019gls}. As shown in Table~\ref{tab:gls}, KGLS-RoCo achieves the lowest average optimality gap on TSP200 and remains highly competitive across other sizes. These results indicate that RoCo can effectively enhance GLS by producing generalizable and high-quality penalty heuristics.

\subsection{Ablation Studies}
\label{sec:ablation}
To better understand the contribution of each core component in the RoCo framework, we conduct a comprehensive ablation study on the TSP dataset under both white-box and black-box prompting scenarios. The ablations target the functional roles of key agents (explorer, exploiter, integrator) as well as the elite mutation mechanism and the multi-agent synergy (MAS), comparing each variant against the full RoCo system and the baseline Evolution of Heuristics (EoH). The results are shown in Table~\ref{tab:ablation}.

Specifically, we disable each role or component individually: the explorer agent responsible for conceptual diversity, the exploiter agent focusing on local refinement, the integrator agent for solution fusion, the elite mutation strategy that leverages historical memory, and the MAS mechanism that coordinates collaborative interactions. Each ablated variant is evaluated over multiple runs, and results are reported as mean ± standard deviation of objective values. Notably, the performance drop under the black-box setting is generally more pronounced, indicating that the absence of any single component hampers the system’s ability to compensate for limited internal feedback. An additional ablation on MAS further shows that while the memory-guided elite mutation alone can outperform the baseline, it still falls short of the full RoCo system. This demonstrates that role-based collaboration provides additional stability and exploration benefits beyond what is achievable through memory alone, especially in black-box scenarios. This underscores the importance of each role and the necessity of structured role-based collaboration in achieving robust heuristic evolution.

We further investigate the effect of the number of collaboration rounds in RoCo. As shown in Table~\ref{tab:ablation}, increasing the number of rounds from 1 to 3 consistently improves performance, particularly under black-box prompting where the gap between single-round and multi-round variants is more pronounced. Extending the ablation to 4 and 5 rounds reveals only marginal gains beyond 3 rounds, confirming that three collaboration rounds achieve an optimal balance between performance and efficiency. This highlights the importance of iterative role interactions for refining heuristic candidates and confirms that multi-round collaboration enhances both robustness and generalization in the RoCo framework.

\section{Conclusion}

This paper presents RoCo, a novel multi-agent role-based system for Automatic Heuristic Design (AHD) that leverages structured collaboration among four specialized LLM-guided agents—explorer, exploiter, critic, and integrator. Integrated into the Evolution of Heuristics (EoH) framework, RoCo enhances heuristic generation through multi-round reasoning, elite mutations, and role-based refinement. Experiments across five combinatorial optimization problems under both white-box and black-box settings demonstrate that RoCo achieves competitive or superior performance compared to state-of-the-art methods. RoCo exhibits faster convergence, better generalization across problem scales, and greater robustness under limited-feedback conditions, setting a new standard for collaborative LLM-based AHD.

Future directions include applying RoCo to more NP-hard problems and other frameworks, beyond combinatorial optimization (e.g., continuous, mixed-integer). Exploring RoCo’s potential in real-world applications, such as logistics and scheduling, remains an exciting avenue for practical impact.

\bibliography{references}

\setcounter{secnumdepth}{1}
\onecolumn
\appendix
\section*{Appendix}
\section{RoCo Algorithm Pseudocode}
\label{sec:roco_pseudocode}

The following pseudocode details the core evolutionary framework of RoCo, including key processes such as population initialization, agent-based heuristic collaboration, memory-guided mutation, and candidate selection for next-generation evolution.

\begin{algorithm}[H]
\caption{RoCo's Evolutionary Framework for Heuristic Population Optimization}
\label{alg:roco}
\begin{algorithmic}[1]
\Require Initial population $P^0$ of size $N$
\For{generation $g = 1$ to $G$}
    \State Apply standard EoH operators (E1--M2) to generate $P_{\text{eoh}}^{g}$
    \State Sample elite pair $(h^{(1)}, h^{(2)}) \subset P^g$
    \State Initial critic comparison between elites:
    \State \quad $(f^{\text{init}}_{\text{cri}}, r^{\text{init}}_{\text{cri}}) = \pi_{\text{critic}}(h^{(1)}, h^{(2)})$
    \State Initialize agent states from $h^{(2)}$ with critic feedback:
    \State \quad $h_0^{\text{explorer}} = \pi_{\text{explorer}}(h^{(2)}, f^{\text{init}}_{\text{cri}})$
    \State \quad $h_0^{\text{exploiter}} = \pi_{\text{exploiter}}(h^{(2)}, f^{\text{init}}_{\text{cri}})$
    \For{round $t = 1$ to $T$}
        \State Critic evaluates current heuristics:
        \State \quad $(f_t^{\text{cri}}, r_t^{\text{cri}}) = \pi_{\text{critic}}(h_{t-1}^{\text{explorer}}, h_{t-1}^{\text{exploiter}})$
        \State Explorer proposes new heuristic:
        \State \quad $h_t^{\text{explorer}} = \pi_{\text{explorer}}(h_{t-1}^{\text{explorer}}, f_t^{\text{cri}})$
        \State Exploiter refines current heuristic:
        \State \quad $h_t^{\text{exploiter}} = \pi_{\text{exploiter}}(h_{t-1}^{\text{exploiter}}, f_t^{\text{cri}})$
        \State Integrator fuses both heuristics:
        \State \quad $h_t^{\text{integrator}} = \pi_{\text{integrator}}(h_t^{\text{explorer}}, h_t^{\text{exploiter}})$
    \EndFor
    \State Identify best heuristics at final round:
    \State \quad $\hat{h}_T^{\text{explorer}} = \text{Best}(h_1^{\text{explorer}}, \dots, h_T^{\text{explorer}})$
    \State \quad $\hat{h}_T^{\text{exploiter}} = \text{Best}(h_1^{\text{exploiter}}, \dots, h_T^{\text{exploiter}})$
    \State Perform elite fusion of best heuristics:
    \State \quad $\hat{h}_T^{\text{integrator}} = \pi_{\text{integrator}}(\hat{h}_T^{\text{explorer}}, \hat{h}_T^{\text{exploiter}})$
    \State Summarize role-specific long-term reflections:
    \State \quad $R_{\text{explorer}}^{\text{lt}} = \text{LTReflect}(R_{\text{explorer}}^{\text{short}}, g_{t-1}, g_t, \Delta g_t)$
    \State \quad $R_{\text{exploiter}}^{\text{lt}} = \text{LTReflect}(R_{\text{exploiter}}^{\text{short}}, g_{t-1}, g_t, \Delta g_t)$
    \State \quad $R_{\text{integrator}}^{\text{lt}} = \text{Merge}(R_{\text{explorer}}^{\text{lt}}, R_{\text{exploiter}}^{\text{lt}})$
    \For{each $h_{\text{base}} \in \{ h^{(1)}, h^{(2)} \}$}
        \State Memory-guided elite mutations:
        \State \quad $h^{\text{mut}}_{\text{exp}} = \pi_{\text{mutation}}^{\text{exp}}(h_{\text{base}}, R_{\text{explorer}}^{\text{lt}})$
        \State \quad $h^{\text{mut}}_{\text{expl}} = \pi_{\text{mutation}}^{\text{expl}}(h_{\text{base}}, R_{\text{exploiter}}^{\text{lt}})$
        \State \quad $h^{\text{mut}}_{\text{int}} = \pi_{\text{mutation}}^{\text{int}}(h_{\text{base}}, R_{\text{integrator}}^{\text{lt}})$
    \EndFor
    \State Construct RoCo candidate set:
    \State \quad $\mathcal{H}^{\text{mut}} = \{ h^{\text{mut}}_{\text{exp}}, h^{\text{mut}}_{\text{expl}}, h^{\text{mut}}_{\text{int}} \}$
    \State \quad $y_{\text{collab}} = \{ h_T^{\text{explorer}}, h_T^{\text{exploiter}}, h_T^{\text{integrator}} \}$
    \State \quad $\mathcal{C}^{g+1}_{\text{RoCo}} = y_{\text{collab}} \cup \mathcal{H}^{\text{mut}} \cup \{ \hat{h}_T^{\text{integrator}} \}$
    \State Merge with standard EoH outputs:
    \State \quad $\mathcal{C}^{g+1}_{\text{full}} = \mathcal{C}^{g+1}_{\text{RoCo}} \cup P_{\text{eoh}}^{g}$
    \State Select top-$N$ heuristics for next generation:
    \State \quad $P^{g+1} = \text{TopN}(\mathcal{C}^{g+1}_{\text{full}}, N)$
\EndFor
\end{algorithmic}
\end{algorithm}

\section{Benchmark Problems}
\subsection{Traveling Salesman Problem}
The Traveling Salesman Problem (TSP) is a classic optimization challenge that seeks the shortest possible route for a salesman to visit each city in a list exactly once and return to the origin city. As one of the most representative combinatorial optimization problems (COPs) , synthetic TSP instances for our experiments are constructed by sampling nodes uniformly from the \([0, 1]^2\) unit square, following the dataset generation protocol described in ReEvo \citep{ye2024reevo}.  

\subsection{Capacitated Vehicle Routing Problem}

The Capacitated Vehicle Routing Problem (CVRP) extends the Traveling Salesman Problem (TSP) by introducing capacity constraints on vehicles. It aims to plan routes for multiple capacity-limited vehicles (starting and ending at a depot) to satisfy customer demands while minimizing total travel distance. For experimental instances: Across experiments, the depot is fixed at the unit square center $(0.5, 0.5)$, following the dataset generation protocol of ReEvo \citep{ye2024reevo}.  

\subsection{Orienteering Problem }
The Orienteering Problem (OP) aims to maximize the total collected score by visiting a subset of nodes within a limited tour length constraint. Following the dataset generation strategy of \citet{ye2024reevo, ye2023deepaco}, we uniformly sample all nodes, including the depot, from the unit square $[0,1]^2$. The prize $p_i$ associated with each node $i$ is set according to a challenging distribution:
\[
p_i = \frac{1 + \frac{j}{99} \cdot \left( \frac{d_{0i}}{\max_{j=1}^{n} d_{0j}} \right)^k}{100},
\]
where $d_{0i}$ denotes the Euclidean distance from the depot to node $i$, and $k$ is a fixed scaling parameter. The tour length constraint is also designed to be challenging, with the maximum length set to 3, 4, 5, 8, and 12 for OP50, OP100, OP200, OP500, and OP1000, respectively.

\subsection{Multiple Knapsack Problem}

The Multiple Knapsack Problem (MKP) involves assigning items (each with a weight and value) to multiple knapsacks to maximize total value, without exceeding any knapsack’s capacity. For instance generation, we follow ReEvo \citep{ye2024reevo}: item values \( v_i \) and weights \( w_{ij} \) (for item \( j \) and knapsack \( i \)) are uniformly sampled from \([0, 1]\), and each knapsack’s capacity \( C_i \) is drawn from \((\max_j w_{ij}, \sum_j w_{ij})\) to ensure valid instances.

\subsection{Offline Bin Packing Problem}
The Offline Bin Packing Problem (Offline BPP) aims to pack a set of items with known sizes into the minimum number of bins of fixed capacity \( W \), where all items are available for assignment upfront. For instance generation, we follow ReEvo \citep{ye2024reevo}: bin capacity is set to \( W = 150 \), and item sizes are uniformly sampled from \([20, 100]\).

\section{Detailed General Framework}

\subsection{Guided Local Search}

Guided Local Search (GLS) explores the solution space via local search operations, where heuristics guide the exploration process. Following the setup in ReEvo \citep{ye2024reevo}, we adopt a modified GLS algorithm that incorporates perturbation phases: edges with higher heuristic values are prioritized for generalization \citep{arnold2019gls}.  

In the training phase, we evaluate candidate heuristics on the TSP200 instance using 1200 GLS iterations. For generating results in our experiments (e.g., Table~\ref{tab:gls}), we use the GLS parameters specified in Table~\ref{tab:gls_params} (consistent with ReEvo). Iterations terminate when a predefined threshold is reached or the optimality gap is reduced to zero.  

\begin{table}[htbp]
\centering
\caption{GLS parameters used for the evaluations in Table 1.}
\label{tab:gls_params}
\begin{tabular}{lccc}  
\hline
\textbf{Problem} & \textbf{Perturbation moves} & \textbf{Number of iterations} & \textbf{Scale parameter }$\boldsymbol{\lambda}$ \\
\hline
TSP20  & 5   & 73   & \multirow{4}{*}{0.1} \\  
TSP50  & 30  & 175  & \\
TSP100 & 40  & 1800 & \\
TSP200 & 40  & 800  & \\
\hline
\end{tabular}
\end{table}

\subsection{Heuristic measures for Ant Colony Optimization}
Ant Colony Optimization (ACO) is a meta-heuristic evolutionary algorithm inspired by ants’ foraging behavior, designed to solve combinatorial optimization problems \citep{dorigo2007aco}. ACO operates via two core matrices:  
\begin{itemize}
    \item Pheromone matrix $\boldsymbol{\tau}$: $\tau_{ij}$ reflects the priority of edge $(i,j)$, iteratively updated based on solution quality to direct searches toward promising areas.  
    \item Heuristic matrix $\boldsymbol{\eta}$: $\eta_{ij}$ encodes problem-specific immediate benefits (e.g., $\eta_{ij} = 1/d_{ij}$ for TSP, where $d_{ij}$ is the inter-city distance).  
\end{itemize}
Ants construct solutions by probabilistically selecting next nodes using a combination of pheromone and heuristic information. A complete ACO iteration includes solution construction, optional local search, and pheromone trail update, enabling convergence to near-optimal solutions for NP-hard problems. Following ReEvo \citep{ye2024reevo}, we use ACO as a baseline framework where LLMs search for optimal heuristics to guide pheromone-informed sampling. For experimental settings in Table \ref{tab:whitebox} and Table \ref{tab:black_box_table}, we follow the protocols defined in MCTS-AHD \citep{zheng2025monte}, which will be shown in Table \ref{tab:acoparameter}.

\begin{table}[htbp]
\captionsetup{width=\linewidth} 
\centering
\caption{ACO parameters for heuristic evaluations. Evaluation phase corresponds to Figure~\ref{fig:whiteboxcurve}, and testing phase corresponds to Table~\ref{tab:whitebox} and Table~\ref{tab:black_box_table}.}
\label{tab:acoparameter}
\begin{tabular}{l|cccc}
\toprule
Problem & Population size & Number of iterations when evaluation & Number of iterations when testing \\
\midrule
TSP  & 30 & 200 & 500 \\
OP   & 30 & 100 & 200 \\
CVRP & 20 & 100 & 500 \\
MKP  & 10 &  50 & 100 \\
BPP  & 20 & 100 & 100 \\
\bottomrule
\end{tabular}
\end{table}

\section{Detailed Evaluations and Experiments}
\label{exp_details}
In this section, we elaborate on the configuration of the evaluation budgets \(T\) and evaluation datasets in different phase. Evaluation settings are generally adopted from early work, including ReEvo \citep{ye2024reevo}, MCTS-AHD \citep{zheng2025monte}.

\paragraph{The settings of \(T\)}
In experiments, all baselines and RoCo have a maximum of 400 evaluations, with a population size of 10 across all datasets. For initial population sizes: ReEvo, HSEvo, and RoCo are 30; Eoh's initial population size matches its population size (consistent with the paper's settings); MCTS is 4, as in the original paper.

\paragraph{The settings of training phase and testing phase evaluation dataset}

In the training and the testing phase, for all tasks, RoCo follows the same settings in LLM-based AHD methods, such as ReEvo \citep{ye2024reevo} and MCTS \citep{zheng2025monte}. For detailed information about the training and testing datasets used across different tasks and frameworks, please refer to Table \ref{tab:dataset}. In the training phase, The running time of each heuristic on the training dataset is limited to 60 seconds except CVRP. The limit time for CVRP is 120 seconds.

\paragraph{White-box and Black-box Settings}

For completeness, we additionally evaluate RoCo and all baselines under both \textbf{white-box} and \textbf{black-box} settings, following recent LLM-based heuristic design protocols.

In the \textbf{white-box setting}, the full problem structure is explicitly exposed to the LLM. For routing problems such as TSP and CVRP, the \emph{distance matrix} is directly included in the prompt, enabling explicit pairwise-distance reasoning. In contrast, the \textbf{black-box setting} restricts structural access: distances are encoded only as abstract edge attributes of shape $(n_{\text{edges}}, 1)$, and prompts cannot access the global structural information. Both prompt templates will be shown in the camera-ready appendix for clarity.









\begin{table}[htbp]
\centering
\caption{Training and testing datasets used for different tasks and frameworks.}
\label{tab:dataset}
\begin{tabular}{c|c|c}
\hline
Framework & ACO & ACO \\
\hline
Task & TSP & OP \\
\hline
Training dataset & 5 50-node TSP instances & 5 50-node OP instances \\
\hline
Testing dataset & 64 50,100,200-node TSP instances & 64 50,100,200,500-node OP instances \\
\hline
Framework & ACO & ACO \\
\hline
Task & CVRP & MKP \\
\hline
Training dataset & 10 50-node CVRP instances & 5 100-item MKP instances (m=5) \\
\hline
Testing dataset & 64 50,100,200,500-node CVRP instances & 64 100,200,300,500,700-item MKP instances \\
\hline
Framework & ACO & GLS \\
\hline
Task & Offline BPP & TSP \\
\hline
Training dataset &5 500-item BPP instances  &10 200-node TSP instances  \\
\hline
Testing dataset & 64 500,700,1000-item BPP instances & 64 50,100,200-node TSP instances \\
\hline
\end{tabular}
\end{table}

\section{Detailed Methodology}

\subsection{Eoh Details}
\label{eohdetails}
In the RoCo system, which is built upon the EoH framework, we adopt four key evolutionary prompt strategies to create new heuristics. These strategies are inspired by human heuristic development behaviors and are categorized into two main types: \textit{Exploration} and \textit{Modification}. The exploration strategies (E1, E2) are designed to diversify the heuristic space, while the modification strategies (M1, M2) aim to refine and improve existing heuristics. Below, we describe each strategy in detail.

\textbf{E1: Dissimilarity-based Exploration.} This strategy encourages the generation of novel heuristics that significantly differ from existing ones. Given a set of $p$ heuristics selected from the current population, the LLM is instructed to design a new heuristic that deviates as much as possible from the selected ones. This helps broaden the search space and introduces innovative ideas that are not present in the current population.

\textbf{E2: Idea-based Exploration.} Unlike E1, this strategy maintains some conceptual consistency with existing heuristics while still aiming for novelty. The LLM is first prompted to identify shared underlying ideas among $p$ selected parent heuristics. It is then asked to generate a new heuristic that builds on these shared principles but introduces new structures or components, thereby maintaining relevance while encouraging creative recombination.

\textbf{M1: Structural Modification.} This strategy directly modifies a single existing heuristic to potentially improve its performance. The LLM is provided with one selected heuristic and prompted to produce a revised version that addresses weaknesses, removes redundancy, or enhances specific parts. This type of local modification supports efficient exploitation in the search space.

\textbf{M2: Parameter Adjustment.} This strategy targets the internal parameters of a given heuristic rather than its overall structure. The LLM is instructed to keep the general logic of the selected heuristic but experiment with different parameter settings. This allows for fine-tuning heuristics while preserving their core behaviors.

These strategies collectively guide the generation of new heuristics in each evolution step. By combining exploration and refinement, the system balances novelty and quality, leading to more effective heuristic discovery over time.

\subsection{Methods for Choosing Elite Individuals from Population}
\label{appendix:elite_selection}

To construct the collaborative pair \( x_{\text{collab}} = (h^{(1)}, h^{(2)}) \), we employ a **probability-based selection strategy** to choose two elite individuals from the current population. The selection process is described as follows:

Given a population of size \( N \), we define a decreasing probability distribution over the individuals, where top-ranked individuals are more likely to be selected. Specifically, the selection probability \( p_i \) for the \( i \)-th individual is:
\[
p_i = \frac{1}{(i+1)^k} \Big/ \sum_{j=0}^{N-1} \frac{1}{(j+1)^k}
\]
where \( k = 3.0 \) is the power coefficient controlling the decay rate of selection preference, favoring top performers while preserving diversity.

The first selected index is sampled from this distribution. The second selected index is then chosen as either the first index + 1 or + 2 (within bounds), ensuring it is greater than the first index and the two individuals are distinct and adjacent in rank. Minimal checks enforce valid ranges, with resampling or defaults applied for edge cases as needed.

This mechanism ensures collaborative pairs are high-quality (elite-biased) and structurally adjacent, promoting meaningful hybridization in subsequent fusion steps.

\subsection{Prompts of RoCo}

\lstdefinestyle{promptstyle}{
  basicstyle=\ttfamily\small,
  backgroundcolor=\color{gray!10},
  frame=single,
  breaklines=true,
  breakatwhitespace=true,
  showstringspaces=false,
  captionpos=b
}

\renewcommand{\lstlistingname}{Prompt}

\begin{lstlisting}[style=promptstyle, caption=The prompt of explorer.]
{task_description}
Here is an algorithm proposed by previous agents:
Algorithm description: {alg_description}
Code: {code}
Objective score: {obj}
Critic's feedback:{cri_response}
Reflection:{reflection}
Based on the critic's feedback and the reflection, please propose a new algorithm that improves the previous ones focusing on global diversity and long-term potential. Use creative, future-oriented thinking to introduce novel directions combined with critic's feedback.
{output_request}
\end{lstlisting}

\begin{lstlisting}[style=promptstyle, caption=The prompt of exploiter.]
{task_description}
Here are some algorithms proposed by previous agents:
Algorithm description: {alg_description}
Code: {code}
Objective score: {obj}
Critic's feedback:{cri_response}
Reflection:{reflection}
Based on the critic's feedback and the reflection, please propose a new algorithm that improves the previous ones to refine and improve it through short-term gains and local optimizations. Use conservative, efficiency-driven thinking to enhance performance incrementally.
{output_request}
\end{lstlisting}

\begin{lstlisting}[style=promptstyle, caption=The prompt of integrator.]
{task_description}
Here are the current candidates for integration:
Explorer's algorithm description: {explorer_algorithm}
Explorer's code:{explorer_code}
Explorer's objective Score: {explorer_obj}

Exploiter's algorithm description: {exploiter_algorithm}
Exploiter's code:{exploiter_code}
Exploiter's objective Score: {exploiter_obj}

Act as a strategic integrator. Weigh the long-term innovation of the explorer, the short-term refinement of the exploiter. Remember A lower objective score is better. Propose the next step that balances exploration and exploitation, maximizing overall progress by reasoning through their complementary strengths.
{output_request}
\end{lstlisting}

\begin{lstlisting}[style=promptstyle, caption=The prompt of final elite integrator.]
{task_description}
Here are the BEST candidates for integration:
Best explorer's code:{explorer_code}
objective Score: {explorer_obj}

Best exploiter's code:{exploiter_code}
objective Score: {exploiter_obj}

Reflection summary: {summary}
Act as a strategic integrator. You must consider:
1. The long-term innovation potential from the best explorer
2. The short-term refinement capabilities from the best exploiter
3. Insights from the reflection summary
Your task is to create a solution that combines the strongest elements from both algorithms while addressing insights from the reflection. Create an algorithm that balances exploration and exploitation to maximize progress.
{output_request}
\end{lstlisting}

\begin{lstlisting}[style=promptstyle, caption=The prompt of initial critic.]
{task_description}
Here are the algorithms proposed by initial population. 
Better code:  
{first_code}  
Better objective score: {first_objective}
Worse code:  
{second_code}  
Worse objective score: {second_objective}
Note: **A lower objective score is better**.

Act as a reflector and critical evaluator in the domain of optimization heuristics. First, reflect on why the better algorithm outperforms the other: You respond with some hints for designing better heuristics, based on the two code and using less than 20 words. Then analyze the worse algorithm's limitations and provide specific, actionable improvements.

Please provide your answer in the following format:  
<ref>Reflection on current code's evolution direction (max 30 words)</ref>
<ans>Specific critique and actionable suggestions (max 40 words)</ans>
Let's think step by step and be constructive in your analysis.
\end{lstlisting}

\begin{lstlisting}[style=promptstyle, caption=The prompt of critic when current individual is better than previous one.]
{task_description}
As an optimization expert in the domain of heuristic optimization, compare:
Worse algorithm description:{first_alg}
Worse and prior code: {first_code}
Worse objective value: {first_obj}
Better algorithm description:{second_alg}
Better and current code: {second_code}
Better objective value: {second_obj}

First, consider and reflect on the evolution direction of the current code:
**Note: current code is worse**: Why did this evolution direction fail? What design choices or implementation changes were counterproductive? Such as "avoid..." and so on. {invalid_ind_prompt}

Then provide targeted feedback for the current code: For the current worse code: Suggest how to correct the evolution path
Format your response as:
<ref>Your reflection about current code</ref> <ans>Specific critique and actionable suggestions (max 40 words)</ans>
Let's think step by step and be constructive in your analysis.
\end{lstlisting}

\begin{lstlisting}[style=promptstyle, caption=The prompt of critic when current individual is worse than previous one.]
{task_description}
As an optimization expert in the domain of heuristic optimization, compare:
Worse algorithm description:{first_alg}
Worse and prior code: {first_code}
Worse objective value: {first_obj}
Better algorithm description:{second_alg}
Better and current code: {second_code}
Better objective value: {second_obj}

First, consider and reflect on the evolution direction of the current code:
**Note: current code is worse**: Why did this evolution direction fail? What design choices or implementation changes were counterproductive? Such as "avoid..." and so on. {invalid_ind_prompt}

Then provide targeted feedback for the current code: For the current worse code: Suggest how to correct the evolution path
Format your response as:
<ref>Your reflection about current code</ref> <ans>Specific critique and actionable suggestions (max 40 words)</ans>
Let's think step by step and be constructive in your analysis.
\end{lstlisting}

\begin{lstlisting}[style=promptstyle, caption=The prompt ofritic when current individual is better than previous one.]
{task_description}
As an optimization expert in the domain of heuristic optimization, compare:
Worse algorithm description:{first_alg}
Worse and prior code: {first_code}
Worse objective value: {first_obj}
Better algorithm description:{second_alg}
Better and current code: {second_code}
Better objective value: {second_obj}

First reflect on the evolution direction of the current code:(Note: Current code is better) What successful evolutionary patterns can we identify? Which design improvements were most effective? Such as "..benefits", "..gains" and so on. {invalid_ind_prompt}

Then provide targeted feedback for the current code: Recommend how to further leverage successful strategies.
Format your response as:
<ref>Your reflection about current code</ref> <ans>Specific critique and actionable suggestions (max 40 words)</ans>
Let's think step by step and be constructive in your analysis.
\end{lstlisting}

\begin{lstlisting}[style=promptstyle, caption=The prompt of long-term reflector.]
Below is your prior short-term reflection on designing heuristics for {task_description}
{reflection}
Based on the relection with changing of the objective scores for {role} {role_description}, write constructive and detailed hints for designing better heuristics, based on prior reflections and focus on {role} and using less than 50 words. Please provide your answer in the following format: <ans>...</ans>. Let's think step by step.
\end{lstlisting}

\begin{lstlisting}[style=promptstyle, caption=The prompt of elite mutation.]
{task_description}
[History reflection as reference]:
{history_reflection}
[Reflection insights]:
{reflection}
[Current best code]:
{elitist_code}
Based on the history reflections and mainly based on reflection insights and the current best code, propose a mutation that addresses the key insights and improves upon the previous solutions.
{output_request}
\end{lstlisting}

\section{Code for Best Heuristics}
\label{sec:best_heuristics_code}

This section presents the best heuristics generated by RoCo for all problem settings, including both white-box and black-box settings.

\lstdefinestyle{codestyle}{
  language=Python,
  basicstyle=\ttfamily\footnotesize,
  keywordstyle=\color{blue},
  commentstyle=\color{gray},
  stringstyle=\color{orange},
  showstringspaces=false,
  frame=single,
  breaklines=true,
  breakatwhitespace=true,
  tabsize=4,
  morekeywords={np},
  captionpos=b
}

\renewcommand{\lstlistingname}{Heuristic}  

\begin{lstlisting}[style=codestyle, caption=RoCo's Best Heuristic for white-box TSP, label=lst:roco_heuristic_white]
def heuristics_v2(distance_matrix):
    num_nodes = distance_matrix.shape[0]
    heuristics_matrix = np.zeros((num_nodes, num_nodes))

    for i in range(num_nodes):
        for j in range(num_nodes):
            if i != j:
                # Modified connection score with increased penalty for high-degree nodes
                degree_penalty = np.sum(distance_matrix[j] < distance_matrix[i]) + 2  # Higher penalty
                distance_score = 1 / (distance_matrix[i][j] ** 3)  # Modified distance scoring
                heuristics_matrix[i][j] = distance_score / degree_penalty

    # Normalize the heuristics matrix
    heuristic_sum = np.sum(heuristics_matrix, axis=1, keepdims=True)
    heuristics_matrix = heuristics_matrix / heuristic_sum

    return heuristics_matrix
\end{lstlisting}

\begin{lstlisting}[style=codestyle, caption=RoCo's Best Heuristic for black-box TSP, label=lst:roco_heuristic_black]
def heuristics_v2(edge_attr: np.ndarray) -> np.ndarray:
    min_attr = np.min(edge_attr)
    max_attr = np.max(edge_attr)
    range_attr = max_attr - min_attr
    
    # Polynomial transformation
    polynomial_attr = np.power(edge_attr - min_attr, 2)  # Using square transformation
    
    # Simplified hybrid scaling technique
    scaled_attr = (polynomial_attr - np.min(polynomial_attr)) / (range_attr + 1e-10)
    
    # Weighted hybrid mean capturing edge attribute diversity
    weighted_mean = np.mean(polynomial_attr)
    heuristics_matrix = (weighted_mean / (scaled_attr + 1e-10)) ** 2  # Prioritize lower attributes
    
    return heuristics_matrix.flatten()  # Return as a 1D array
\end{lstlisting}

\begin{lstlisting}[style=codestyle, caption=RoCo's Best Heuristic for white-box OP]
def heuristics(prize, distance, maxlen):
    n = len(prize)
    heuristics_matrix = np.zeros((n, n))

    for i in range(n):
        prize_distance_ratios = []
        for j in range(n):
            if i != j and distance[i][j] <= maxlen:
                ratio = prize[j] / distance[i][j]
                exponential_ratio = np.exp(prize[j] / distance[i][j])
                combined_score = 0.5 * ratio + 0.5 * exponential_ratio  # Equal weight to both ratios
                prize_distance_ratios.append((j, combined_score))

        # Sort nodes by the combined score in descending order
        prize_distance_ratios.sort(key=lambda x: x[1], reverse=True)
        
        # Calculate cumulative promise while respecting maxlen
        accumulated_distance = 0
        for j, score in prize_distance_ratios:
            if accumulated_distance + distance[i][j] <= maxlen:
                heuristics_matrix[i][j] = score
                accumulated_distance += distance[i][j]

    return heuristics_matrix
\end{lstlisting}

\begin{lstlisting}[style=codestyle, caption=RoCo's Best Heuristic for black-box OP]
def heuristics_v2(node_attr, edge_attr, node_constraint):
    n = len(node_attr)
    heuristics_matrix = np.zeros((n, n))

    special_node_attr = node_attr[0]  # Attribute of the special node indexed by 0

    # Initialize adaptive thresholds based on a new factor
    edge_threshold = np.maximum(edge_attr.sum(axis=0) * 0.7, node_constraint)  # New threshold adjustment factor
    score_history = np.zeros(n)  # to store history of scores for feedback mechanism
    fixed_weight = 0.25  # Adjusted fixed weight for scoring

    for i in range(1, n):
        for j in range(n):
            if i != j:  # Skip self-loops
                edge_sum = edge_attr[i, j]
                if edge_sum > 0 and edge_sum <= edge_threshold[j]:
                    connection_diversity = np.sum(edge_attr[i] > 0)
                    diversity_contribution = (special_node_attr * node_attr[j] * connection_diversity) / (edge_sum ** 2 + 1e-10)

                    # Introduce modified fixed weight into score calculation
                    weighted_score = fixed_weight * (node_attr[j] ** 2) / (edge_sum + 1e-10)
                    score = diversity_contribution * (weighted_score * (node_constraint / (edge_sum + 1e-10)))

                    # Update thresholds based on previous scores with adjustments for sensitivity
                    edge_threshold[j] = max(edge_threshold[j], score / (1 + score_history[j] ** 0.5))  # dynamic adjustment with sensitivity

                    # Ensure non-negativity and log scores for feedback mechanism
                    heuristics_matrix[i, j] = max(0, score)
                    score_history[j] = score  # Store the latest score

    return heuristics_matrix
\end{lstlisting}

\begin{lstlisting}[style=codestyle,caption=RoCo's Best Heuristic for white-box CVRP]
def heuristic_v2(distance_matrix, coordinates, demands, capacity):
    n = distance_matrix.shape[0]
    heuristics_matrix = np.zeros((n, n))

    historical_performance = np.ones((n, n))
    total_edge_visits = np.zeros((n, n))

    learning_rate = 0.3
    regularization_term = 0.01

    for i in range(1, n):
        for j in range(1, n):
            if i != j:
                total_demand = demands[i] + demands[j]

                distance_score = 1 / (distance_matrix[i, j] ** 2 + 1e-5)  # Reduced emphasis on proximity
                proximity_score = 1 / (np.linalg.norm(coordinates[i] - coordinates[j]) ** 2 + 1e-5)

                if total_demand <= capacity:
                    edge_score = distance_score * proximity_score * historical_performance[i, j]
                else:
                    dynamic_penalty = (total_demand ** 2) / (capacity + 1e-5)  # Heavier penalty for high demand
                    edge_score = (distance_score * proximity_score * historical_performance[i, j]) / dynamic_penalty

                # Adjusting clustering factor to consider only higher than average demand
                avg_demand = np.mean(demands[1:])
                clustering_factor = sum(demands[k] for k in range(1, n) if demands[k] > avg_demand and np.linalg.norm(coordinates[k] - coordinates[i]) < np.linalg.norm(coordinates[k] - coordinates[j])) / max(avg_demand, 1e-5)
                edge_score *= (1 + clustering_factor)

                decay_factor = 1 / (1 + total_edge_visits[i, j])
                historical_performance[i, j] = (1 - learning_rate) * historical_performance[i, j] + learning_rate * edge_score
                historical_performance[i, j] = historical_performance[i, j] / (1 + regularization_term * total_edge_visits[i, j])

                total_edge_visits[i, j] += 1

                heuristics_matrix[i, j] = edge_score

    return heuristics_matrix

\end{lstlisting}

\begin{lstlisting}[style=codestyle, caption=RoCo's Best Heuristic for black-box CVRP]
def heuristics(edge_attr, node_attr):
    import numpy as np

    n = edge_attr.shape[0]
    heuristics_matrix = np.zeros_like(edge_attr)

    normalization_factor = np.power(node_attr, 1/3)
    adaptive_weights = np.ones_like(edge_attr)

    max_iterations = 100
    tolerance = 1e-5
    exploration_factor = 0.1
    damping_factor = 0.8  # Damping factor to mitigate oscillations

    previous_heuristics_matrix = np.copy(heuristics_matrix)

    for iteration in range(max_iterations):
        for i in range(n):
            for j in range(n):
                if edge_attr[i, j] != 0:
                    heuristics_matrix[i, j] = (normalization_factor[i] * normalization_factor[j]) / (edge_attr[i, j] ** 2)
                    adaptive_weights[i, j] = damping_factor * adaptive_weights[i, j] + (1 - damping_factor) * heuristics_matrix[i, j]

        heuristics_matrix *= adaptive_weights
        
        # Calculate mean and std deviation for adaptive weight adjustments
        positive_heuristics = heuristics_matrix[heuristics_matrix > 0]
        if len(positive_heuristics) > 0:
            mean_heuristics = np.mean(positive_heuristics)
            std_dev = np.std(positive_heuristics)

            for i in range(n):
                for j in range(n):
                    if edge_attr[i, j] != 0:
                        adaptive_weights[i, j] = np.clip(adaptive_weights[i, j] + 0.01 * (mean_heuristics - std_dev), 0, None)

        if np.max(np.abs(heuristics_matrix - previous_heuristics_matrix)) < tolerance:
            break

        previous_heuristics_matrix = np.copy(heuristics_matrix)

    return heuristics_matrix
\end{lstlisting}

\begin{lstlisting}[style=codestyle,caption=RoCo's Best Heuristic for white-box MKP]
def heuristics_reevo(prize, weight):
    n = len(prize)
    m = weight.shape[1]
    heuristics_matrix = np.zeros(n)

    global_average = np.mean(prize)
    performance_metrics = np.zeros(n)  # Track performance of selections
    selection_counts = np.zeros(n)  # Track the number of selections made
    temperature = 1.0  # Initial temperature for exploration
    max_iterations = 100  # Maximum iterations for the process
    iteration = 0  # Current iteration
    reset_threshold = 20  # Periodic reset interval
    synergy_matrix = np.zeros((n, n))  # Store synergy potential between items

    while iteration < max_iterations:
        for i in range(n):
            max_weight = np.max(weight[i])
            if max_weight > 0:
                efficiency_score = prize[i] / max_weight  # Efficiency of prize-to-weight ratio
                diversity_factor = 1 / (1 + np.abs(prize[i] - global_average))  # Encourages diversity

                # Update performance metrics
                if selection_counts[i] > 0:
                    performance_metrics[i] = (performance_metrics[i] * 0.5 + efficiency_score * 0.5)
                else:
                    performance_metrics[i] = efficiency_score

                # Synergy potential computation with diversity weighting
                for j in range(n):
                    if i != j:
                        synergy_weight = np.exp(-np.abs(weight[i] - weight[j]).sum())  # High synergy if weights are similar
                        synergy_matrix[i][j] = synergy_weight * (1 + 0.5 * np.abs(prize[i] - prize[j]) / (np.max(prize) - np.min(prize)))        

                # Calculate total synergy score with normalization
                total_synergy = np.sum(synergy_matrix[i]) / (n - 1)  # Normalize
                heuristics_matrix[i] = efficiency_score * diversity_factor * performance_metrics[i] * total_synergy

                # Dynamic exploration-exploitation balancing
                adaptive_exploration_probability = 1 / (1 + selection_counts[i] * temperature)
                if np.random.rand() < adaptive_exploration_probability:
                    selection_counts[i] += 1

        # Periodic exploration reset
        if iteration % reset_threshold == 0 and iteration > 0:
            performance_metrics = np.zeros(n)  # Reset performance metrics to enhance exploration

        # Temperature decay adjustment
        average_performance = np.mean(performance_metrics) + 0.1  # Update based on performance
        temperature = max(0.1, temperature * (0.99 + 0.01 * average_performance))  # Gradually decay temperature

        iteration += 1

    return heuristics_matrix

\end{lstlisting}

\begin{lstlisting}[style=codestyle, caption=RoCo's Best Heuristic for black-box MKP]
def heuristics_v2(item_attr1, item_attr2):
    n = len(item_attr1)
    m = item_attr2.shape[1]
    total_weight = np.sum(item_attr2, axis=1, keepdims=True) + 1e-6  # Avoid division by zero
    heuristics_matrix = np.zeros(n)

    for i in range(n):
        max_weight = np.max(item_attr2[i]) + 1e-6
        normalized_ratio = (item_attr1[i] / max_weight) / total_weight[i]  # Normalize with total weight
        item_contributions = np.sum(item_attr2[i])
        
        # Safety check to prevent numerical instability
        if item_contributions > 0:
            heuristics_matrix[i] = np.power(normalized_ratio, 2) * item_contributions  # Weight by item contributions

    heuristics_matrix = np.clip(heuristics_matrix, 0, None)  # Ensure non-negative values
    return heuristics_matrix
\end{lstlisting}

\begin{lstlisting}[style=codestyle, caption=RoCo's Best Heuristic for white-box Offline BPP]
def heuristics_v2(demand, capacity):
    n = len(demand)
    heuristics_matrix = np.zeros((n, n))

    total_demand = np.sum(demand)
    sorted_indices = np.argsort(demand)[::-1]
    bins = []

    # First-fit decreasing approach
    for idx in sorted_indices:
        placed = False
        for b in bins:
            if sum(demand[i] for i in b) + demand[idx] <= capacity:
                b.append(idx)
                placed = True
                break
        if not placed:
            bins.append([idx])

    # Inter-cluster optimization with different scoring
    for b in bins:
        cluster_size = len(b)
        for i in range(cluster_size):
            for j in range(i + 1, cluster_size):
                idx1 = b[i]
                idx2 = b[j]
                if demand[idx1] + demand[idx2] <= capacity:
                    used_capacity = demand[idx1] + demand[idx2]
                    average_item_size = (demand[idx1] + demand[idx2]) / 2
                    score = (1 / (1 + (capacity - used_capacity) + (average_item_size / capacity)))
                    heuristics_matrix[idx1][idx2] = heuristics_matrix[idx2][idx1] = score

    return heuristics_matrix
\end{lstlisting}

\begin{lstlisting}[style=codestyle, caption=RoCo's Best Heuristic for black-box Offline BPP]
def heuristics(node_attr: np.ndarray, node_constraint: int) -> np.ndarray:
    n = node_attr.shape[0]
    heuristics_matrix = np.zeros((n, n))

    num_clusters = min(n // 2, 5)
    kmeans = KMeans(n_clusters=num_clusters)
    clusters = kmeans.fit_predict(node_attr.reshape(-1, 1))

    for i in range(n):
        for j in range(n):
            if i != j:
                combined_attr = node_attr[i] + node_attr[j]
                if combined_attr <= node_constraint:
                    cluster_score = 1 if clusters[i] == clusters[j] else 0.5
                    attractiveness_score = np.exp(-np.abs(node_constraint - combined_attr))
                    heuristics_matrix[i, j] = attractiveness_score * cluster_score
                else:
                    heuristics_matrix[i, j] = 0

    return heuristics_matrix

\end{lstlisting}

\begin{lstlisting}[style=codestyle, caption=RoCo's Best Heuristic for TSP GLS]
def heuristics_v2(distance_matrix):
    num_nodes = distance_matrix.shape[0]
    heuristics_matrix = np.zeros_like(distance_matrix)

    for i in range(num_nodes):
        for j in range(num_nodes):
            if i != j:
                # Calculate the heuristic value for each edge (i, j)
                heuristics_matrix[i][j] = (distance_matrix[i][j] / (np.sum(distance_matrix[i]) + np.sum(distance_matrix[j])))

    return heuristics_matrix

\end{lstlisting}

\end{document}